\documentclass{article}
\usepackage{spconf,amsmath,amssymb,graphicx,booktabs,multirow}
\usepackage[T1]{fontenc}
\usepackage[utf8]{inputenc}
\usepackage{microtype}
\usepackage{float}
\makeatletter
\long\def\@makecaption#1#2{%
  \vskip 1.5pt
  \setbox\@tempboxa\hbox{#1. #2}%
  \ifdim \wd\@tempboxa >\hsize
    #1. #2\par
  \else
    \hbox to\hsize{\hfil\box\@tempboxa\hfil}%
  \fi}
\renewcommand\section{\@startsection{section}{1}{\z@}%
  {-1.8ex plus -.4ex minus -.2ex}{0.65ex plus .2ex}{}}
\renewcommand\subsection{\@startsection{subsection}{2}{\z@}%
  {-1.5ex plus -.3ex minus -.2ex}{0.5ex plus .2ex}{}}
\makeatother
\title{E-AVI: Evidence-Grounded Multimodal Assessment for Automated Video Interviews}
\name{\begin{tabular}{c}
Haoshen Wang$^{1,\star}$ \qquad Dongbo Che$^{1,\star}$ \qquad Zeyi Xie$^{2}$ \\
Yuanjie Du$^{1}$ \qquad Shicheng Hua$^{1}$ \qquad Xingyu Wang$^{1}$
\end{tabular}}
\address{$^{1}$The Hong Kong Polytechnic University \qquad
$^{2}$Independent Researcher \qquad $^{\star}$Equal contribution}
\begin{document}
\setlength{\abovedisplayskip}{5pt plus 2pt minus 2pt}
\setlength{\belowdisplayskip}{5pt plus 2pt minus 2pt}
\setlength{\abovedisplayshortskip}{3pt plus 1pt minus 1pt}
\setlength{\belowdisplayshortskip}{4pt plus 1pt minus 1pt}
\maketitle
\begin{abstract}
Automated video interview assessment integrates verbal content, acoustic delivery, and visual behavior, yet numerical predictions alone provide limited inspectable support. We present E-AVI, an evidence-grounded framework that extracts timestamped multimodal evidence and integrates dimension-conditioned evidence attention with source-level embeddings for scoring. A shared evidence pool further supports natural-language feedback and follow-up question answering. On RecruitView and a private hospitality dataset, E-AVI consistently outperforms fine-tuned multimodal baselines in rank correlation. Ablation, evidence-deletion, bootstrap, human-audit, and QA analyses characterize the predictive contribution, grounding, and practical utility of the evidence pathway. Together, these results demonstrate that our proposed E-AVI framework improves predictive performance while providing inspectable support for assessment, feedback, and interactive analysis.
\end{abstract}

\section{Introduction}

Automated video interviews (AVIs) predict interview-related ratings from recorded responses at scale~\cite{hickman2022automated,hickman2025smart,chen2017automated}. By modeling verbal content, acoustic delivery, and visual behavior, they assess candidate performance efficiently~\cite{naim2016automated,chen2016automatic,koutsoumpis2024beyond}. However, most AVI systems remain score-centric: they output numerical predictions while providing limited insight into the observations underlying each assessment.

This limitation is increasingly important as multimodal models are used for prediction, interpretation, and interaction. Prior work has studied the psychometric properties of AVI assessment~\cite{liff2024psychometric}, as well as fine-grained interpretation and fairness-oriented modeling~\cite{rahman2021hirepreter,putra2024mag}. Recent multimodal models can express behavioral observations in natural language and generate detailed explanations. Yet fluent explanations are not necessarily grounded in the source recording, nor do they necessarily reflect the evidence that contributes to scoring~\cite{atanasova2023faithfulness}. A useful assessment system should therefore provide evidence that is both inspectable and connected to the predictive process, particularly given the importance of transparency for applicant trust in AI-based AVIs~\cite{suen2023building}.

We present E-AVI, an evidence-grounded framework for multimodal interview assessment. Instead of directly mapping an interview to scores alone, E-AVI first extracts structured, timestamped evidence from textual, acoustic, and visual signals. A dimension-conditioned scorer then attends to evidence relevant to each assessment dimension and combines it with source-level multimodal embeddings for prediction. The resulting evidence pool serves as a shared intermediate representation for natural-language feedback and follow-up QA, enabling users to inspect and query assessment evidence.

We evaluate E-AVI on the public RecruitView dataset~\cite{gupta2025recruitview} and a private hospitality interview dataset collected in real-world interview settings. Across both datasets, E-AVI consistently improves predictive performance over fine-tuned multimodal baselines. We further examine whether the extracted evidence is grounded in the source recordings and whether it meaningfully contributes to prediction through human evidence auditing, component ablations, evidence deletion, uncertainty analysis, difficult-case analysis and QA evaluation.

Our main contributions are:
\begin{itemize}
    \setlength{\itemsep}{0pt}
    \setlength{\parsep}{0pt}
    \setlength{\topsep}{2pt}
    \setlength{\partopsep}{0pt}
    \item An evidence-grounded framework for multimodal interview assessment with structured, timestamped evidence.
    \item Evaluation on public and private interview datasets, with consistent improvements in predictive performance over fine-tuned multimodal baselines.
    \item Analyses of evidence grounding, predictive contribution, statistical uncertainty, difficult cases, and QA quality.
\end{itemize}
\section{Method}
\begin{figure*}[t]
\centering
\includegraphics[width=\textwidth]{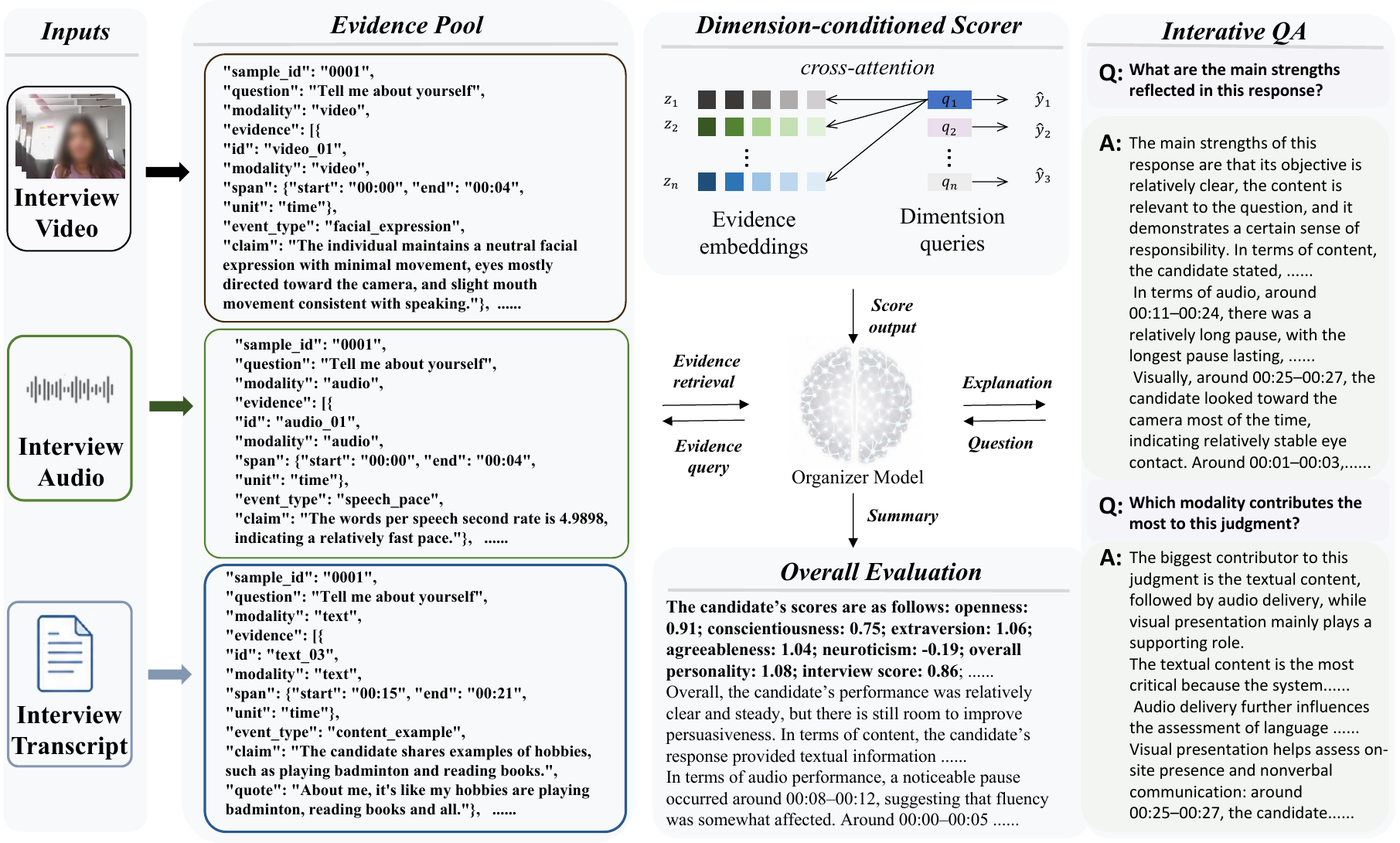}
\caption{Overview of E-AVI. Source embeddings complement the explicit evidence pathway during scoring.}
\label{fig:overview}
\end{figure*}
\subsection{Evidence extraction and corpus construction}
Given video, audio, and transcript inputs, an evidence extractor produces $E=\{e_i\}_{i=1}^{N}$. Each evidence item contains an identifier, modality, event type, temporal span, and natural-language claim, as illustrated in Fig.~\ref{fig:overview}. We train MiniCPM-o 4.5~\cite{cui2026minicpm} as a unified evidence extractor using teacher-generated supervision. Specifically, Qwen3-VL-32B-Instruct~\cite{bai2025qwen3}, Qwen3.5, and Qwen3.5-Omni~\cite{team2026qwen3} serve as modality-specific teachers for visual, transcript, and acoustic evidence, respectively. Deterministic descriptors of speech activity, pauses, pitch, energy, and speaking rate further supplement the audio input with explicit acoustic cues complementary to learned representations.

For each video, the modality-specific teachers first generate an initial evidence pool under the shared schema. These teacher-generated evidence items supervise the extractor to map multimodal interview inputs into structured, timestamped observations.

To support follow-up QA, we train the extractor in a question-conditioned recheck mode. Given a question, the multimodal input, and the current evidence pool, recheck searches for question-relevant observations missed during initial extraction or returns no new evidence. It is invoked at inference when the existing pool is insufficient for answering a follow-up question.

We construct recheck supervision from RecruitView training samples using automatically generated follow-up questions rather than human QA annotations. In Mask-and-Ask, one evidence item is withheld from the current pool, and a modality-specific teacher generates a third-party evaluator question that requires the withheld observation and cannot be answered from the remaining evidence. In Second-Pass Mining, modality-specific teachers or evidence-derived templates construct questions from the complete evidence pool and original modality input. These questions cover recoverable observations omitted from the pool, queries already supported by existing evidence, and unsupported queries for which no new evidence should be returned. The corresponding target is either newly recovered evidence or an explicit no-new-evidence outcome.

\subsection{Dimension-conditioned scoring}
Each evidence item is serialized from its modality, event type, temporal span, and natural-language claim and mapped to a cached embedding $\mathbf h_i$. Source embeddings are obtained by running the LoRA-adapted MiniCPM-o extractor on the full multimodal input, interview question, and transcript, then mean-pooling final-layer LLM states over visual, acoustic, and textual token spans to yield $\mathbf u_v$, $\mathbf u_a$, and $\mathbf u_t$. They provide contextual information complementary to explicit evidence. Evidence and source tokens are projected separately and combined as attention memory:
\begin{equation}
\begin{aligned}
\mathbf z_i&=\operatorname{LN}(W_e\mathbf h_i), &
\mathbf g_m&=\operatorname{LN}(W_s\mathbf u_m)+\mathbf t_m,\\
\mathbf M&=[\mathbf z_1;\ldots;\mathbf z_N;\mathbf g_v;\mathbf g_a;\mathbf g_t].
\end{aligned}
\end{equation}
where $\mathbf t_m$ is a learned source-type embedding. For dimension $d$, a learned query attends jointly to all evidence and source tokens:
\begin{equation}
\begin{aligned}
(\mathbf r_d,\boldsymbol\alpha_d)&=\operatorname{MHA}(\mathbf q_d,\mathbf M,\mathbf M),\\
\hat y_d&=f_d\!\left(\operatorname{LN}(\mathbf q_d+\operatorname{Dropout}(\mathbf r_d))\right).
\end{aligned}
\end{equation}
Here $\boldsymbol\alpha_d$ contains attention over both token types. Its evidence portion is renormalized as $\bar\alpha_{d,i}=\alpha_{d,i}/\sum_{j=1}^{N}\alpha_{d,j}$ and retained with evidence identifiers as dimension-specific support; source attention represents complementary information and is not treated as readable evidence.

Weak attention supervision uses rule-derived relevance targets $R_{d,i}$ based on modality constraints and event-type mappings, such as speech pace to speaking skills and content structure to answer quality. Let $\mathcal P_d=\{(i,j):R_{d,i}>R_{d,j}\}$ and $w_{d,ij}=R_{d,i}-R_{d,j}$. Training minimizes
\begin{equation}
\mathcal L=\mathcal L_{\rm Huber}+\lambda_{\rm attn}
\frac{\sum_d\sum_{(i,j)\in\mathcal P_d}w_{d,ij}[m-\bar\alpha_{d,i}+\bar\alpha_{d,j}]_+}
{\sum_d\sum_{(i,j)\in\mathcal P_d}w_{d,ij}}.
\end{equation}
This objective supervises only evidence attention, leaving source-token attention unconstrained. We use predefined scoring dimensions because natural-language rubrics are unavailable for the public dataset.

\begin{table}[!t]
\centering
\small
\caption{RecruitView results under three settings.}
\label{tab:results_recruit}
\renewcommand{\arraystretch}{0.90}
\begin{tabular*}{\linewidth}{@{\extracolsep{\fill}}lrrrrrr@{}}
\toprule
Method & MAE$\downarrow$ & MSE$\downarrow$ & $\rho\uparrow$ & $\tau\uparrow$ & C$\uparrow$ & r$\uparrow$ \\
\midrule
\multicolumn{7}{@{}l}{\textit{Zero-shot}} \\
VL2 & 0.886 & 1.527 & -0.061 & -0.041 & 0.479 & -0.043 \\
VL2-AV & 0.779 & 1.288 & 0.038 & 0.027 & 0.514 & 0.041 \\
Q3-VL & 0.982 & 1.740 & -0.359 & -0.260 & 0.378 & -0.264 \\
GPT-4o & 0.964 & 1.706 & -0.420 & -0.301 & 0.357 & -0.334 \\
MCPM-o & 0.826 & 1.366 & -0.163 & -0.114 & 0.445 & -0.110 \\
\addlinespace[1pt]
\multicolumn{7}{@{}l}{\textit{Few-shot}} \\
VL2 & 0.904 & 1.562 & -0.080 & -0.053 & 0.473 & -0.024 \\
VL2-AV & 0.878 & 1.579 & -0.050 & -0.034 & 0.483 & -0.022 \\
Q3-VL & 1.032 & 1.943 & -0.229 & -0.161 & 0.420 & -0.136 \\
GPT-4o & 1.010 & 1.830 & -0.341 & -0.234 & 0.384 & -0.242 \\
MCPM-o & 1.001 & 2.397 & -0.206 & -0.151 & 0.431 & -0.179 \\
\addlinespace[1pt]
\multicolumn{7}{@{}l}{\textit{Fine-tuned}} \\
VL2 & 0.693 & 1.228 & 0.381 & 0.262 & 0.630 & 0.318 \\
VL2-AV & 0.740 & 1.264 & 0.347 & 0.240 & 0.620 & 0.263 \\
Q3-VL & 0.691 & 1.025 & 0.373 & 0.255 & 0.627 & 0.319 \\
MCPM-o & 0.641 & 0.930 & 0.440 & 0.307 & 0.654 & 0.402 \\
\textbf{E-AVI} & \textbf{0.607} & \textbf{0.895} & \textbf{0.541} & \textbf{0.385} & \textbf{0.693} & \textbf{0.492} \\
\bottomrule
\end{tabular*}
\end{table}

\begin{table}[!t]
\centering
\small
\caption{Private hospitality results under three settings.}
\label{tab:results_private}
\renewcommand{\arraystretch}{0.90}
\begin{tabular*}{\linewidth}{@{\extracolsep{\fill}}lrrrrrr@{}}
\toprule
Method & MAE$\downarrow$ & MSE$\downarrow$ & $\rho\uparrow$ & $\tau\uparrow$ & C$\uparrow$ & r$\uparrow$ \\
\midrule
\multicolumn{7}{@{}l}{\textit{Zero-shot}} \\
VL2-AV & 1.357 & 1.964 & 0.119 & 0.231 & 0.530 & 0.190 \\
VL2 & 1.123 & 1.722 & 0.364 & 0.339 & 0.642 & 0.330 \\
Q3-VL & 1.100 & 1.750 & 0.188 & 0.092 & 0.537 & 0.183 \\
GPT-4o & 1.314 & 2.279 & 0.328 & 0.357 & 0.631 & 0.306 \\
MCPM-o & 0.671 & 0.700 & 0.305 & 0.346 & 0.614 & 0.399 \\
\addlinespace[1pt]
\multicolumn{7}{@{}l}{\textit{Few-shot}} \\
VL2-AV & 1.121 & 1.746 & 0.196 & 0.235 & 0.570 & 0.209 \\
VL2 & 0.929 & 1.386 & 0.281 & 0.319 & 0.606 & 0.327 \\
Q3-VL & 0.671 & 0.821 & 0.256 & 0.276 & 0.601 & 0.289 \\
GPT-4o & 1.279 & 2.154 & 0.270 & 0.296 & 0.603 & 0.328 \\
MCPM-o & 0.707 & 0.768 & 0.393 & 0.482 & 0.639 & 0.463 \\
\addlinespace[1pt]
\multicolumn{7}{@{}l}{\textit{Fine-tuned}} \\
VL2 & 0.585 & 0.665 & 0.425 & 0.382 & 0.664 & 0.410 \\
VL2-AV & 0.536 & 0.589 & 0.481 & 0.506 & 0.687 & 0.471 \\
Q3-VL & 0.623 & 0.636 & 0.445 & 0.375 & 0.688 & 0.377 \\
MCPM-o & 0.661 & 0.701 & 0.328 & 0.262 & 0.631 & 0.398 \\
\textbf{E-AVI} & \textbf{0.503} & \textbf{0.506} & \textbf{0.639} & \textbf{0.542} & \textbf{0.771} & \textbf{0.595} \\
\bottomrule
\end{tabular*}
\end{table}

\subsection{Feedback and follow-up QA}
At inference, scoring is performed once and the predicted scores and evidence pool are cached. GPT-4o~\cite{hurst2024gpt} serves as an organizer for feedback and follow-up QA. For each question, it retrieves relevant evidence and answers directly when sufficient; otherwise, it invokes question-conditioned recheck with the question, relevant multimodal context, and current evidence pool. The extractor then adds targeted observations or returns no new evidence, while cached scores remain unchanged.

\section{Experiments}
\subsection{Datasets and comparison protocol}
RecruitView~\cite{gupta2025recruitview} contains 2,011 response videos from over 300 participants answering 76 interview questions, with video, audio, transcripts, and 12-dimensional continuous ratings. We use participant-level 80\%/10\%/10\% training/validation/test splits to avoid identity overlap. The private dataset consists of structured interviews with 101 frontline employees from three hotels in southern China. Each participant answers five questions, with 60 seconds for preparation and 60 seconds for response. Question-level HR ratings range from 1 to 5 in 0.5-point increments. We use the same participant-level split proportions. The extractor trained on RecruitView is transferred to the private dataset without further fine-tuning, while only the scorer is trained on the private training split.

Baselines are VideoLLaMA2 and its audio-visual variant~\cite{cheng2024videollama}, Qwen3-VL-8B, GPT-4o, and MiniCPM-o-9B. In few-shot evaluation, all methods receive the same three videos, randomly sampled from the corresponding training split. Zero/few-shot prompts request fixed-order numeric outputs: 12 values for RecruitView and one for the private dataset. Text is parsed to floats in that order; missing or non-numeric outputs are regenerated, while out-of-range numeric values remain unchanged. Because E-AVI uses teacher-generated evidence supervision, these are complete-system rather than matched-supervision architecture comparisons. Tables~\ref{tab:results_recruit} and~\ref{tab:results_private} abbreviate the baselines as VL2, VL2-AV, Q3-VL, and MCPM-o. We report MAE, MSE, Spearman's $\rho$, Kendall's $\tau$, C-index, and Pearson's $r$.

\subsection{Implementation details}
Extractor supervision uses only RecruitView~\cite{gupta2025recruitview} training videos. MiniCPM-o 4.5 is LoRA-adapted in bfloat16 ($r=16$, $\alpha=32$, dropout 0.05; maximum length 8,192)~\cite{hu2021lora}; the final four-epoch mixed initial/recheck stage uses batch size 1, gradient accumulation 8, and AdamW with learning rate $2\times10^{-5}$. Invalid, incomplete, or unparsable JSON is regenerated.

Serialized evidence items are encoded separately by mean-pooling the final-layer MiniCPM-o LLM states. Both evidence and source embeddings are cached as 4,096-dimensional vectors. The scorer projects them to 256 dimensions and uses four attention heads, dropout 0.1, and a two-layer GELU head per target. It is trained for 100 epochs with batch size 24, AdamW ($8\times10^{-4}$ learning rate, $10^{-4}$ weight decay), $\lambda_{\rm attn}=0.06$, $m=0.05$, gradient clipping at 1.0, and seed 42, with checkpoint selection by lowest validation MAE.

With models preloaded and initialization excluded, batch-1 latency averages 15 s for extraction, 0.01 s for scoring, 3 s for QA, and 18 s with recheck, using one GPU with approximately 28 GB of memory. These measurements are setup-specific.

\subsection{Scoring performance and uncertainty}
On RecruitView, E-AVI reduces MAE from 0.641 to 0.607 and improves $\rho$ from 0.440 to 0.541 over fine-tuned MiniCPM-o-9B. On private interviews, MAE is 0.503 and $\rho=0.639$, with the extractor transferred unchanged.

For the small private test set, we bootstrap participants with replacement for 1,000 iterations, retaining every selected participant's responses. E-AVI's 95\% intervals are [0.47,0.59] for MAE, [0.43,0.71] for $\rho$, and [0.73,0.85] for C-index. Paired resampling against VideoLLaMA2-AV gives an MAE reduction of 0.033, with 95\% interval [0.021,0.041]. This supports an improvement on the current test split; it does not establish stability across new splits, hotels, or populations.

\subsection{Evidence contribution and difficult-case analysis}
Table~\ref{tab:ablation} separates modality and component effects. Removing text gives the largest modality-related MAE increase. Removing evidence changes MAE only from 0.607 to 0.612, but reduces $\rho$ from 0.541 to 0.435. Removing source embeddings causes a much larger correlation loss to 0.261. Weak attention supervision improves rank consistency, while its removal slightly improves MAE. Explicit evidence therefore benefits ranking, but is neither the sole predictive source nor uniformly optimal for every metric.

\begin{table}[!htbp]
\centering
\caption{RecruitView ablations. WAS: weak attention supervision.}
\label{tab:ablation}
\small
\setlength{\tabcolsep}{3pt}
\begin{tabular}{lrrrr}
\toprule
Variant & MAE & $\rho$ & C-index & Pearson\\
\midrule
Full E-AVI & 0.607 & 0.541 & 0.693 & 0.492\\
No text & 0.650 & 0.439 & 0.652 & 0.390\\
No audio & 0.613 & 0.458 & 0.662 & 0.428\\
No video & 0.615 & 0.490 & 0.672 & 0.447\\
\midrule
No evidence & 0.612 & 0.435 & 0.652 & 0.380\\
No source emb. & 0.670 & 0.261 & 0.588 & 0.223\\
No WAS & 0.601 & 0.515 & 0.683 & 0.476\\
\bottomrule
\end{tabular}
\end{table}

\begin{figure}[!htbp]
\centering
\includegraphics[width=\columnwidth]{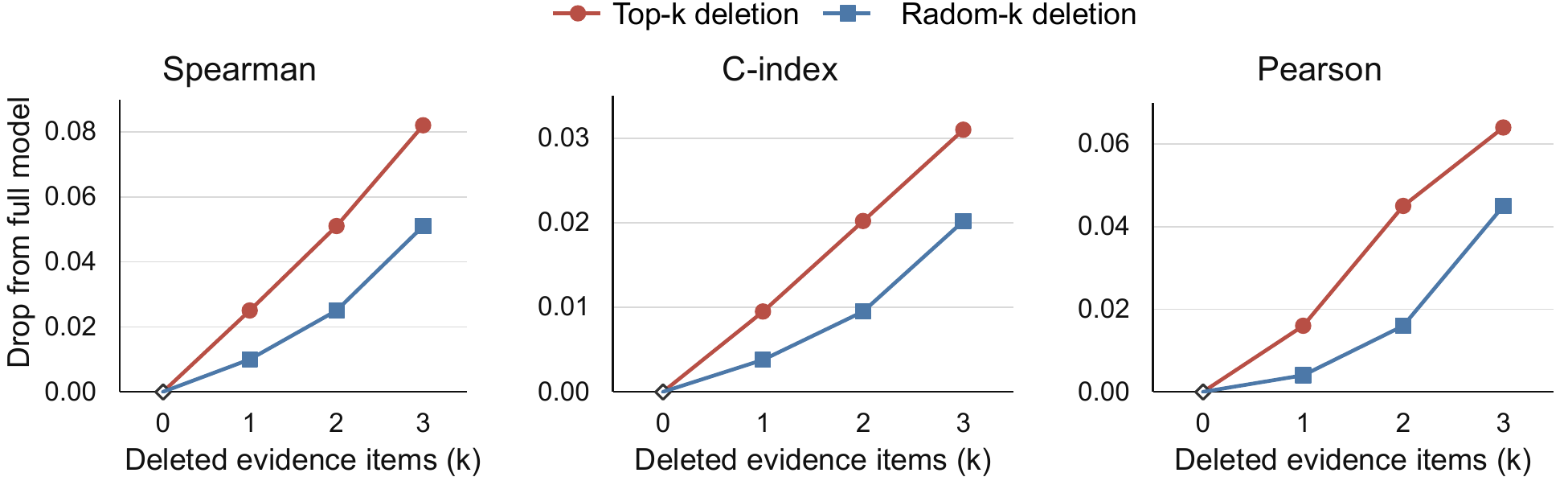}
\caption{Performance drops after deleting top-attended versus random evidence with source embeddings retained.}
\label{fig:faith}
\end{figure}

Deleting the top-$k$ attended evidence items produces larger performance drops than random-$k$ deletion (Fig.~\ref{fig:faith}), particularly for Spearman and Pearson correlation, with increasing gaps as $k$ grows. This tests predictive sensitivity to selected evidence, not complete causal attribution. Its moderate effect is consistent with the parallel source-embedding pathway and redundant observations.

We inspect the 30 samples with the largest residuals between E-AVI and fine-tuned MiniCPM-o-9B, where the residual is defined as E-AVI MAE minus baseline MAE, averaged over 12 dimensions. These represent the cases most challenging for E-AVI relative to the baseline: E-AVI underperforms on 19 samples and only marginally outperforms on the remaining 11. Manual inspection identifies the dominant limiting factor as extraction error (17/30), missing evidence (8/30), or scoring error (5/30). Overall, the evidence-extraction stage is implicated in 25/30 (83.3\%) cases, suggesting that evidence correctness and completeness remain the main bottleneck.

\subsection{Human evidence audit}
\label{sec:quality}
Two annotators inspect 870 evidence items from 50 randomly selected videos against the corresponding source content. Each item is categorized as fully supported, partially supported/overclaimed, or unsupported, and Table~\ref{tab:audit} reports the averaged percentages across annotators. Textual evidence achieves the highest support rate, while 80.8\% of all evidence items are fully supported overall. This audit directly assesses whether the extracted natural-language evidence is grounded in the original recordings and complements the preceding predictive analyses. It evaluates the factual support of generated claims rather than evidence recall or downstream QA quality.

\begin{table}[!htbp]
\centering
\caption{Human evidence audit (\%). Partial includes overclaims; percentages may not sum to 100 due to rounding.}
\label{tab:audit}
\small
\begin{tabular}{lrrrr}
\toprule
Modality & Items & Full & Partial & Unsupported\\
\midrule
Text & 305 & 89.5 & 5.9 & 4.5\\
Audio & 290 & 74.8 & 17.9 & 7.2\\
Video & 275 & 77.4 & 12.7 & 9.8\\
Overall & 870 & 80.8 & 12.1 & 7.1\\
\bottomrule
\end{tabular}
\end{table}

\subsection{Human evaluation of QA}
\begin{figure}[H]
\centering
\includegraphics[width=\columnwidth]{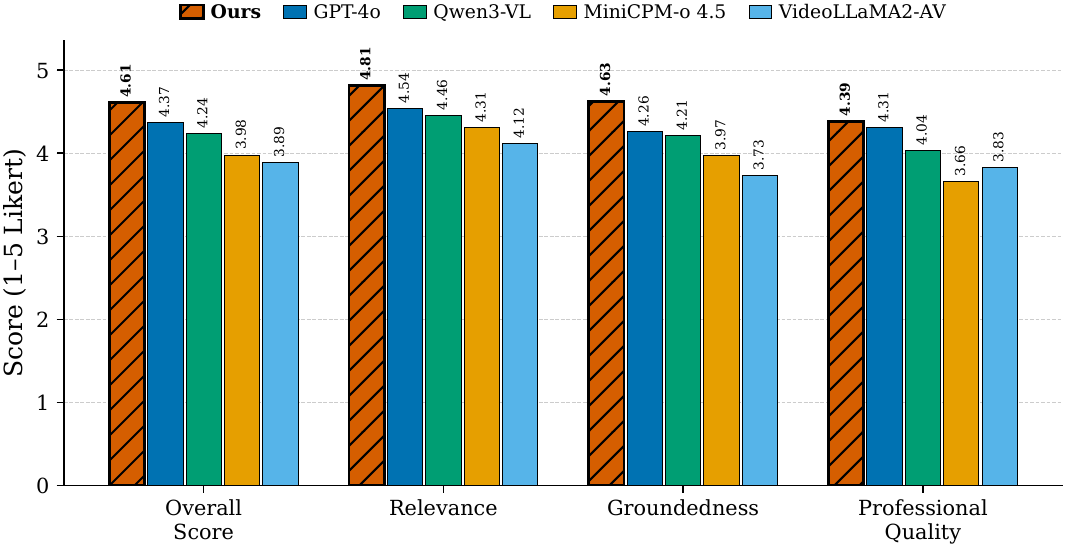}
\caption{Human QA ratings for the evaluated systems (1--5 scale). These compare complete pipelines rather than matched-supervision explanation mechanisms.}
\label{fig:qa}
\end{figure}
Two senior researchers rate answers to eight questions for 50 random RecruitView videos (1--5). Against GPT-4o, Qwen3-VL, MiniCPM-o, and VideoLLaMA2-AV, E-AVI ranks highest (Fig.~\ref{fig:qa}): 4.609 overall, 4.814 relevance, 4.626 groundedness, and 4.386 professional quality. Corresponding average-rater absolute-agreement ICCs are 0.769, 0.683, 0.644, and 0.795; agreement is lowest for groundedness. Repeated-measures analysis shows significant overall method effects across all criteria $p<0.001$.
\section{Conclusion}
E-AVI presents an evidence-grounded framework that connects multimodal interview scoring, feedback, and follow-up QA through a shared pool of structured, timestamped evidence. Across both RecruitView and the private hospitality dataset, E-AVI consistently improves predictive performance over fine-tuned multimodal baselines. Ablation and evidence-deletion analyses further show that explicit evidence contributes meaningfully to prediction and complements source-level multimodal representations. Human auditing demonstrates that most extracted evidence is supported by the original recordings, while the QA evaluation highlights the utility of the evidence pool for grounded and interactive assessment. Together, these results suggest that making multimodal evidence an explicit intermediate representation can improve both predictive performance and the inspectability of automated interview assessment, providing a practical foundation for more transparent and interactive AVI systems.

\clearpage

\bibliographystyle{IEEEbib}
\bibliography{custom}
\end{document}